\documentclass[letterpaper, 10 pt, conference]{ieeeconf}  % Comment this line out if you need a4paper
\usepackage{amsmath}
\usepackage{amsfonts}
\usepackage{graphicx}
\usepackage{xcolor}
\usepackage{multirow}
\usepackage{booktabs}
\usepackage{wrapfig}
\usepackage{adjustbox}
\usepackage{float}
\usepackage{hyperref}
\usepackage{svg}
\usepackage{subfig}

\usepackage{etoolbox}
\makeatletter
\patchcmd{\@makecaption}
  {\scshape}
  {}
  {}
  {}
\makeatother

\newcommand{\Reals}{\mathbb{R}}

\DeclareMathOperator*{\argmin}{arg\,min}

\IEEEoverridecommandlockouts                              % This command is only needed if 
\title{\LARGE \bf
TACO: Trajectory-Aware Controller Optimization for Quadrotors}

\author{Hersh Sanghvi$^{1 \dagger}$, Spencer Folk$^1$, Vijay Kumar$^1$, Camillo Jose Taylor$^1$ % <-this % stops a space
\thanks{$^{1}$School of Engineering and Applied Science, University of Pennsylvania}
\thanks{$^{\dagger}$Corresponding Author (email: hsanghvi@seas.upenn.edu)}
}

\begin{document}

\maketitle
\thispagestyle{empty}
\pagestyle{empty}

%%%%%%%%%%%%%%%%%%%%%%%%%%%%%%%%%%%%%%%%%%%%%%%%%%%%%%%%%%%%%%%%%%%%%%%%%%%%%%%%
\begin{abstract}

Controller performance in quadrotor trajectory tracking depends heavily on parameter tuning, yet standard approaches often rely on fixed, manually tuned parameters that sacrifice task-specific performance. We present Trajectory-Aware Controller Optimization (TACO), a framework that adapts controller parameters online based on the upcoming reference trajectory and current quadrotor state. TACO employs a learned predictive model and a lightweight optimization scheme to optimize controller gains in real time with respect to a broad class of trajectories, and can also be used to adapt trajectories to improve dynamic feasibility while respecting smoothness constraints. To enable large-scale training, we also introduce a parallelized quadrotor simulator supporting fast data collection on diverse trajectories. Experiments on a variety of trajectory types show that TACO outperforms conventional, static parameter tuning while operating orders of magnitude faster than black-box optimization baselines, enabling practical real-time deployment on a physical quadrotor. Furthermore, we show that adapting trajectories using TACO significantly reduces the tracking error obtained by the quadrotor.

\end{abstract}

%%%%%%%%%%%%%%%%%%%%%%%%%%%%%%%%%%%%%%%%%%%%%%%%%%%%%%%%%%%%%%%%%%%%%%%%%%%%%%%%
\section{Introduction}
The performance of many control algorithms is often dependent on the choice of parameters that govern their behavior. A useful property of classical and optimization-based controllers, which are widespread in robotics, is that they can be easily reconfigured for performance on new tasks by changing their parameters, such as feedback gains or cost weights. However, this also presents a drawback as these parameters must often be carefully tuned for the controller to work well. This often means that there is a tradeoff between peak performance on an individual task and average performance across a variety of tasks. Recent work has shown that adapting controllers to the demands of specific tasks can yield significant performance improvements \cite{loquercioAutoTuneControllerTuning2022, kunapuli2025leveling}, though manually retuning for each individual task is too burdensome to be practical. 

Much of the prior research on controller optimization has focused on mitigating model uncertainty and improving robustness under unmodeled dynamics in a reactive fashion. By contrast, the question of how to tune control parameters for optimal performance on an \textbf{upcoming} reference trajectory segment remains relatively underexplored. This perspective is particularly relevant for trajectory tracking in aerial robots, where the difficulty of the tracking task depends not only on the system dynamics but also on the structure of the commanded trajectory. 

In this work, we address this gap by investigating how control parameters for a quadrotor UAV can be automatically tuned with respect to the upcoming reference trajectory and the current state of the robot. Our contributions are threefold: 
\begin{enumerate}
    \item We present a novel framework for adapting controller parameters conditioned on the upcoming reference trajectory using a learned predictive model. 
    \item We demonstrate how this predictive model can also be used to optimize the reference trajectories to increase their dynamic feasibility while also respecting trajectory constraints.
    \item We introduce an accelerated version of an open source aerial robot simulator to facilitate fast, large-scale data collection and model training.
\end{enumerate}

\begin{figure}[t]
\centering
\includegraphics[width=0.45\textwidth]{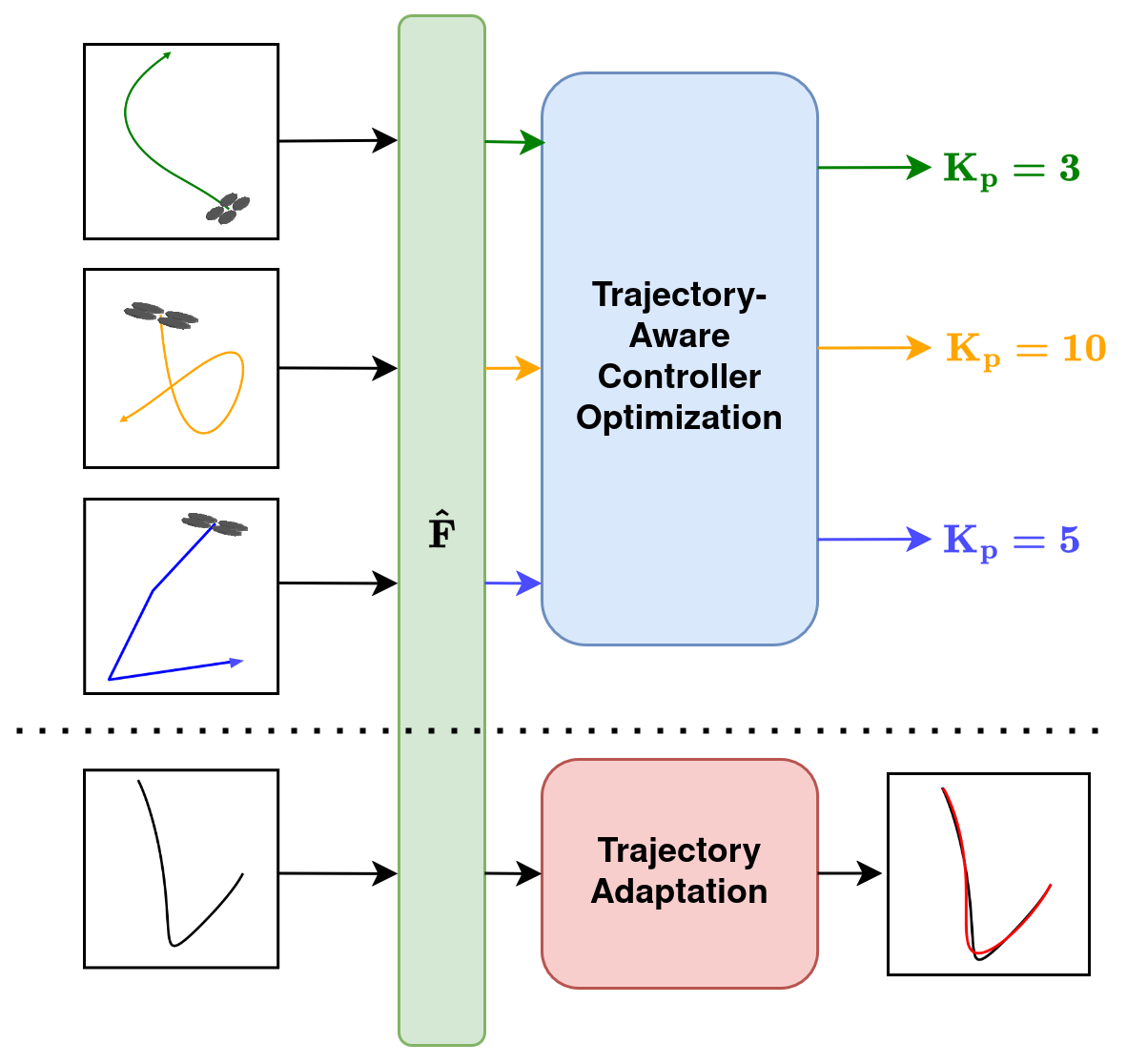}
\caption{We train a model to predict a quadrotor controller's performance based on the upcoming reference trajectory, and demonstrate that this model can be used to optimize the control parameters in real-time for different upcoming trajectory segments. We also use the same predictive model to adapt reference trajectories to increase dynamic feasibility.}
\label{fig:headline}
\end{figure}

\section{Related Work}

Classical heuristic-based PID tuning approaches (e.g. Ziegler-Nichols) remain the most popular approach to gain tuning. These are usually offline procedures based on the controller's step or impulse response. However, heuristic- or hand-based tuning requires domain expertise and does not scale well with the number of tuning parameters. Using control theoretic methods to adjust a control system based on reference trajectory dynamics with stability guarantees has been explored in methods such as Model Reference Adaptive Control (MRAC) \cite{6282874}, L1-Adaptive Control \cite{wuL1Adaptive2022}, and online gain scheduling \cite{RUGH20001401}, though these require accurate characterization of the plant dynamics or "rules of thumb" as to how the parameters should be changed (e.g. the filter bandwidth in L1-Adaptive control or design of the scheduling variables in gain scheduling).

Towards the removal of heuristics and improving scalability, controller adaptation and optimization for trajectory tracking with quadrotors is a well-explored problem, with a variety of methods having been proposed. Offline approaches include variants of Bayesian Optimization \cite{berkenkampSafeControllerOptimization2016} and Monte-Carlo sampling \cite{loquercioAutoTuneControllerTuning2022}. Online approaches include auto-differentiation through the closed-loop system dynamics \cite{chengDiffTuneAutoTuningAutoDifferentiation2024, pan2025learningflyrapidpolicy} and modeling controller performance using meta-learning \cite{pmlr-v270-sanghvi25a, richardsAdaptiveControlOrientedMetaLearningNonlinear2021a, oconnellNeuralFly2022}. Many of the online approaches focus on using \textbf{past} data to estimate disturbances, as opposed to tuning specifically for the \textbf{upcoming} task.  

Meanwhile, some recent works have suggested that controller performance can be further improved by tuning the controller differently depending on the shape of the reference trajectory. \cite{loquercioAutoTuneControllerTuning2022} use a heuristic to divide the reference trajectory into segments based on the altitude gradient and find parameters for an MPC controller for each segment individually. \cite{kunapuli2025leveling} use offline Bayesian Optimization to tune the gains of a geometric controller for different trajectory types and show favorable comparisons to a reinforcement learning-based tracking controller. Providing controllers with a future trajectory horizon is also common in RL tracking controllers \cite{huang2023datt, kaufmann_benchmark_rl, kunapuli2025leveling} and Model Predictive Control (MPC) \cite{data_driven_mpc}. Also related to our work, \cite{srikanthan2025quadlcdlayeredcontroldecomposition} present a method to adapt the \textit{reference trajectory} ahead of the quadrotor to increase its dynamic feasibility to prevent motor saturation effects. We combine aspects of the above approaches to train a single model that can both optimize the controller parameters and the upcoming reference trajectory in an \textbf{online}, receding-horizon fashion.

\section{Problem Setup}
We consider the problem of finding the optimal parameters for a quadrotor controller to track an upcoming reference trajectory based on the current state of the quadrotor. 

Given a control policy $C$ with controller parameters $g$ (e.g. feedback gains or cost weights) that produces control inputs $u$ to track a reference trajectory $\tau$, we seek to solve the following optimization problem of finding the optimal gains over a horizon $H$:
\begin{align}
    g^* &= \argmin_g J(c) \nonumber \\
    &= \argmin_g J(F_g(o_{n:n+H}, u_{n:n+H}, \tau_{n:n+H})) \label{eq:cost}
\end{align}

The cost function of the optimization problem is a scalarization $J(\cdot)$ of some set of cumulative \textit{performance measures} $c$ over the horizon $H$. $n$ denotes the current discretized time, with an associated timestep $\delta t$: $t=n\delta t$, and $H$ denotes the number of discrete timesteps in the horizon.
The scalarization function $J$ is typically a weighted sum of the terms in $c$: $J(c) = w^T c$ for some $w$. These performance measures consist of any terms the designer wishes to optimize, such as tracking error, stability, and control effort. We refer to the function that maps the robot states $o$, robot actions $u$, and commanded trajectory $\tau$ to $c$ as the \textit{performance function} $F$. Because the closed-loop dynamics are a function of the control parameters $g$, $F$ is also implicitly a function of $g$.

\subsection{Quadrotor Controller}
As a case study, in this work we examine a geometric tracking controller for quadrotors \cite{leeGeometric2010}, which we restate here for convenience. Informally, the controller computes a feedforward thrust and moment based on a reference trajectory and nominal model parameters, and adds additional thrusts and torques using feedback on the error between the current actual and desired states. Denoting the current three-dimensional position, velocity, attitude, and angular velocity tracking errors respectively as  $e_p, e_v, e_R, e_{\Omega} \in \Reals ^3$, the commanded thrust and moment are computed as: 
\begin{align*}
    f &= (-k_pe_p -k_v e_v -mge_3 + m\ddot{p}_d)Re_3 \\
    M &= -k_Re_R - k_{\Omega}e_{\Omega} + (\Omega \times J\Omega) \\
    & - J (\hat{\Omega}R^TR_d\Omega_d - R^TR_d\dot{\Omega}_d)
\end{align*}

Where $e_3$ represents the direction of the $z$ axis in the inertial frame and $m,J$ are the quadrotor's mass and inertia matrix. The gains $k_p, k_v \in \Reals^3$ and $k_R, k_{\Omega} \in \Reals$ collectively form the 8-dimensional parameters of the controller, that is $g = [k_p^\top, k_v^\top, k_R, k_\Omega]^\top \in \Reals^8$. In simulation, the commanded thrust and moment are mapped to individual rotor thrusts using a control allocation matrix, from which the motor speeds are computed using the estimated thrust coefficient. For more details on the mathematical derivation and properties of this controller, we refer readers to \cite{leeGeometric2010}. Although this controller is exponentially stable in theory, it does not explicitly take into account effects such as rotor drag, motor lag and control saturation, as well as the dynamic feasibility of the commanded trajectory. Usually, the feedback gains are tuned to compensate for these effects.
\begin{figure*}[t] % use * for spanning both columns
    
    \centering
    \subfloat[The model that predicts performance measures based on control gains, upcoming trajectory, and current quadrotor state.\label{fig:predictive_model}]{
        \includegraphics[width=0.45\textwidth]{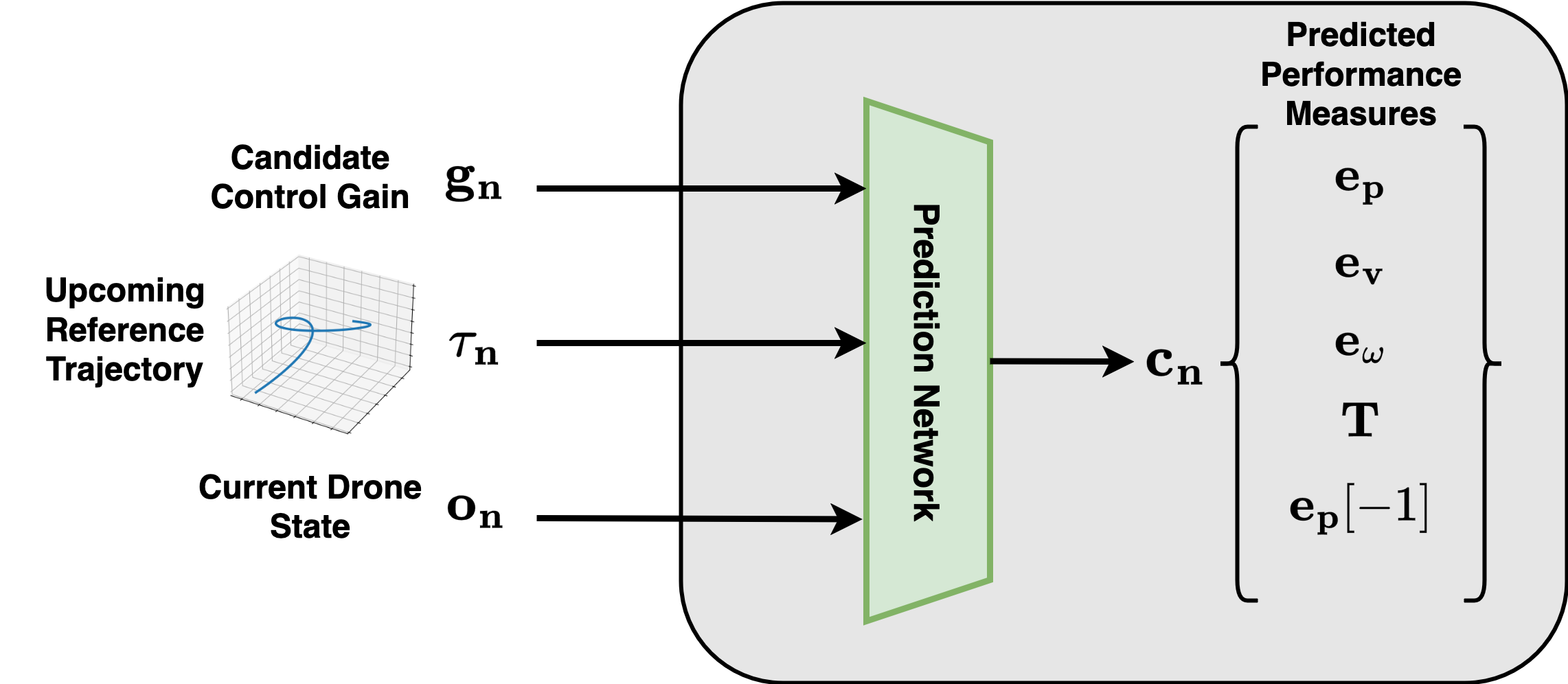}
    }
    \hfill
    \subfloat[Online controller optimization with the predictive model. At regular intervals $T$, TACO observes the upcoming trajectory and the quadrotor's current state, optimizes the control parameters, and deploys those parameters. \label{fig:gain_opt}]{
        \includegraphics[width=0.48\textwidth]{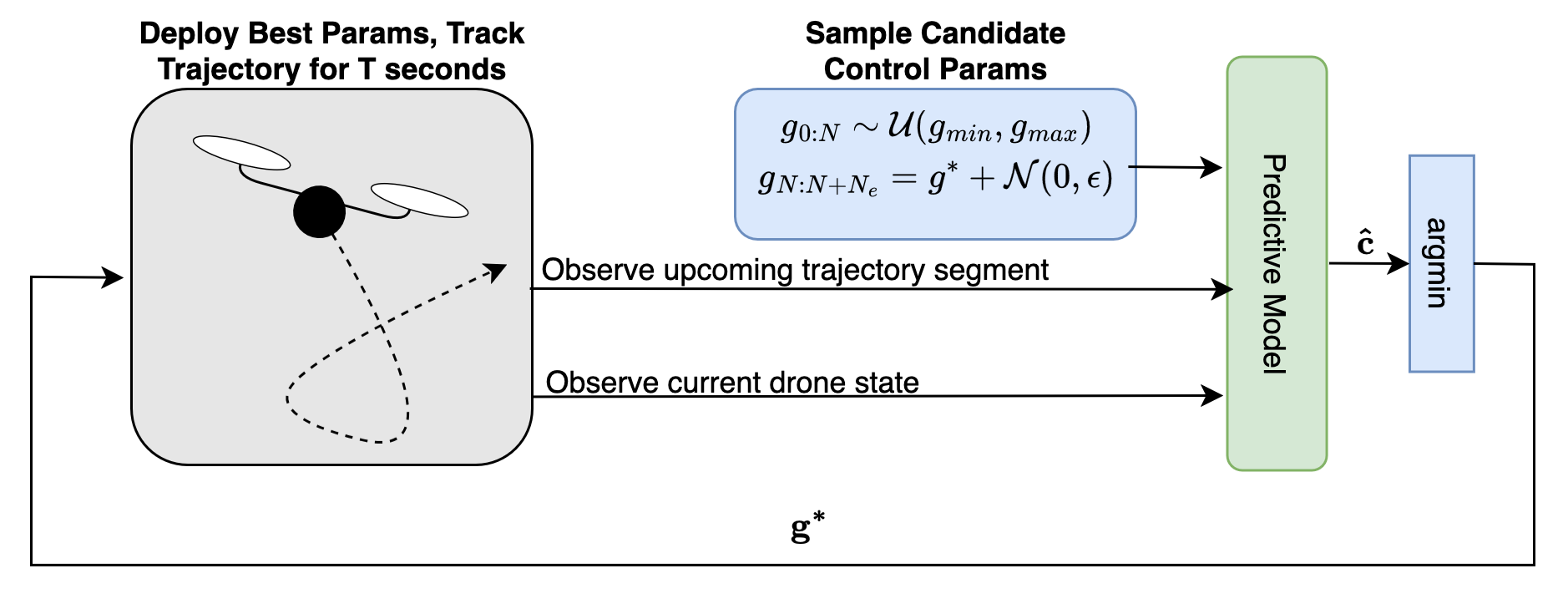}
    }
    \caption{TACO's predictive model (left) and online gain optimization procedure (right)}
    \label{fig:model_and_gain_opt}
\end{figure*}

\subsection{Case Study: Trajectory-specific Gain Tuning}
\begin{table}
\vspace{0.5em}
\centering
\caption{Tracking Error (m) achieved in cross-comparisons of control parameters on four different trajectories. Parameters are tuned independently for each trajectory and tested on all other trajectories. On each test trajectory, the best-performing parameters were the ones that were tuned specifically for that trajectory.}
\label{tab:optuna_crossval}
\resizebox{0.4\textwidth}{!}{%
\begin{tabular}{lllllll}
    &                          & \multicolumn{4}{l}{\;\;\;\;\;\;\;\;\;Tested on} \\ \cline{3-6}
\multirow{5}{*}{\rotatebox[origin=c]{90}{Tuned For}} &                          & Slow       & Med        & Fast      & X-Fast   \\ \cline{3-6} 
    & \multicolumn{1}{l|}{Slow} & \textbf{0.012}     & 0.074      & 0.46      & 0.75  \\
    & \multicolumn{1}{l|}{Med}   & 0.04     & \textbf{0.07}     & 0.38      & 0.69  \\
    & \multicolumn{1}{l|}{Fast} & 0.092     & 0.14     & \textbf{0.2}     & 0.29  \\
    & \multicolumn{1}{l|}{X-Fast}   & 0.087     & 0.13      & 0.2      & \textbf{0.28}
\end{tabular}}
\end{table}

As a simple experiment to establish the benefits of trajectory-specific gain tuning, we use Bayesian Optimization (implemented in the Optuna library \cite{optuna_2019}) to tune the control gains of the geometric controller to track four reference trajectories in simulation, each with similar shape but increasing velocity. We then cross-validate the tuned gains found for each trajectory on the other trajectories and report the results in Table \ref{tab:optuna_crossval}. For all four trajectories, the best-performing control gains were the ones specifically found for that particular trajectory, showing that there are significant performance gains that can be achieved by tuning the controller for a particular trajectory.

\section{Method}

\subsection{Predicting Controller Performance}
The main component of our method is a learned model $\hat{F}$, modeled as a multilayer perceptron that approximates the controller performance function given a candidate set of control gains $g = [k_p^x, k_p^y, k_p^z, k_v^x, k_v^y, k_v^z, k_R, k_\Omega] \in \Reals ^8$, an observation of the quadrotor state at timestep $n$, and the upcoming reference trajectory. Critically, this paper extends the predictor in \cite{pmlr-v270-sanghvi25a} by also conditioning it on a the trajectory lookahead $\bar{\tau}_{n:n+H}$ (which is omitted in their quadrotor experiments due to tuning only on simplistic trajectories). Over the upcoming horizon $H$, the reference trajectory is discretized at regular intervals $\Delta t = m\delta t$ (where $m$ is some positive integer), the desired positions are flattened into a vector and shifted relative to the current position of the quadrotor:
\begin{align*}
    \tau(t) &= [x_d(t), y_d(t), z_d(t)] \\
    \bar{\tau}_{n:n+H} &= [\tau(n),\; \tau(n+m), \cdots, \; \tau(n+H)] - p_n
\end{align*}

The observation of the current robot state, $o_n = \left[v_n, q_n, \omega_n\right] \in \Reals^{10}$ consists of the quadrotor's current velocity $v_n$, quaternion representing the rotation from body frame to inertial frame $q_n$, and current angular velocity in the body frame $\omega_n$.

These inputs are concatenated together and fed into the model, which outputs the predicted performance measures $c_{n:n+H}$. In this work, $c$ is an 8-dimensional vector containing the: 
\begin{enumerate}
\item average absolute tracking errors in the $x$, $y$, $z$ axes: $\frac{1}{H}\sum_{i=n}^{n+H} |e_{p}^{\{x, y, z\}}[i]|$
\item Average velocity tracking error: $\frac{1}{H} \sum_{i=n}^{n+H} ||e_{v}[i]||$
\item Average absolute pitch/roll velocity: $\frac{1}{H} \sum_{i=n}^{n+H} ||\omega[i]||$
\item Average control effort (in terms of commanded thrust and moment): $\frac{1}{H} \sum_{i=n}^{n+H} T_i + M_i$
\item The terminal tracking error, or the difference between the desired and actual position at the end of the lookahead horizon $H$, similar to a terminal cost used to maintain persistent feasibility in receding-horizon control: $e_p[n+H]$
\end{enumerate}

A diagram of our predictive model is shown in Figure \ref{fig:predictive_model}.
\subsection{Controller Optimization}
At regular intervals $T \leq H$, the predictive model is used to find optimal control gains for the upcoming segment of the reference trajectory. Substituting our model's predicted cost into the ground-truth cost in equation \ref{eq:cost} yields the following optimization problem:
\begin{align}
    \hat{g}^*_n &= \argmin_{g_n} J(\hat{F}(g_n, o_n, \bar{\tau}_{n:n+H}))
\end{align}

In the case of the quadrotor's geometric tracking controller, the optimization problem over the controller gains is relatively low-dimensional, so an iterative random search algorithm is used to approximately solve the optimization problem, similar to \cite{pmlr-v270-sanghvi25a}. Effectively, this approach uses importance sampling to explore both random regions in the gain space and those with high potential. Random search is also highly parallelizable with a neural network; in each search iteration, all search samples can be batched and evaluated with a single forward pass through the network. In the first search iteration, we generate $N$ random search samples from a uniform distribution $G_r = \mathcal{U}(g_{\text{min}}, g_{\text{max}})$. Each sample is concatenated with the upcoming trajectory, $\bar{\tau}$ and current state $o_t$ and passed through the network to obtain the expected cost. After evaluating all samples, the sample with the lowest expected cost is chosen. In the following search iterations, we apply $N_r$ random perturbations to the optimal gain vectors found in the previous iterations, generating $N_r$ additional samples $G_e$, and again generate $N$ random search samples to evaluate through the network. A diagram of the online gain optimization is shown in Figure \ref{fig:gain_opt}.

\subsection{Trajectory Optimization} \label{sec:traj_adaptation}
In the previous section, we used the predictive model only to optimize the control gains. 
However, the predictive model $\hat{F}$ can also be used to optimize the \textit{reference trajectories} to make them more dynamically feasible for the quadrotor to track. In this case, the desired downstream task is for the quadrotor to pass through a set of keypoints $p_i$ at times $t_i$, with the reference trajectory encoding a smooth path for the quadrotor to follow between the keypoints. We specifically consider trajectories consisting of smooth, polynomial splines that connect the keypoints in space \cite{mellingerMinimumSnapTrajectory2011}. For polynomial spline segments of the form:
\begin{align*}
    \tau_i(t) &= \sigma_0 + \sigma_1t + \sigma_2t^2 + ... + \sigma_d t^d 
\end{align*}
the full set of polynomial spline coefficients can be found as the solution to a linear system of equations:
\begin{align*}
    A \left[ \sigma_0^x, \; \sigma_0^y, \; \sigma_0^z, \; \sigma_1^x, ... \sigma_{n_s}^z \right]^T &= b
\end{align*}
where $\sigma_i \in \Reals^{d+1}$ denotes the polynomial coefficients of the $i$th $d$-degree polynomial spline segment. $A,b$ encode constraints such as:
\begin{enumerate}
    \item Each spline segment $i$ must begin at keypoint $p_i$ at time $t_i$ and end at keypoint $p_{i+1}$ at time $t_{i+1}$.
    \item The velocity, acceleration, and jerk must be continuous between adjacent spline segments at each keypoint (ie. at intersections of the segments)
\end{enumerate}

\begin{figure}[t]
    \centering
    \includegraphics[width=0.45\textwidth]{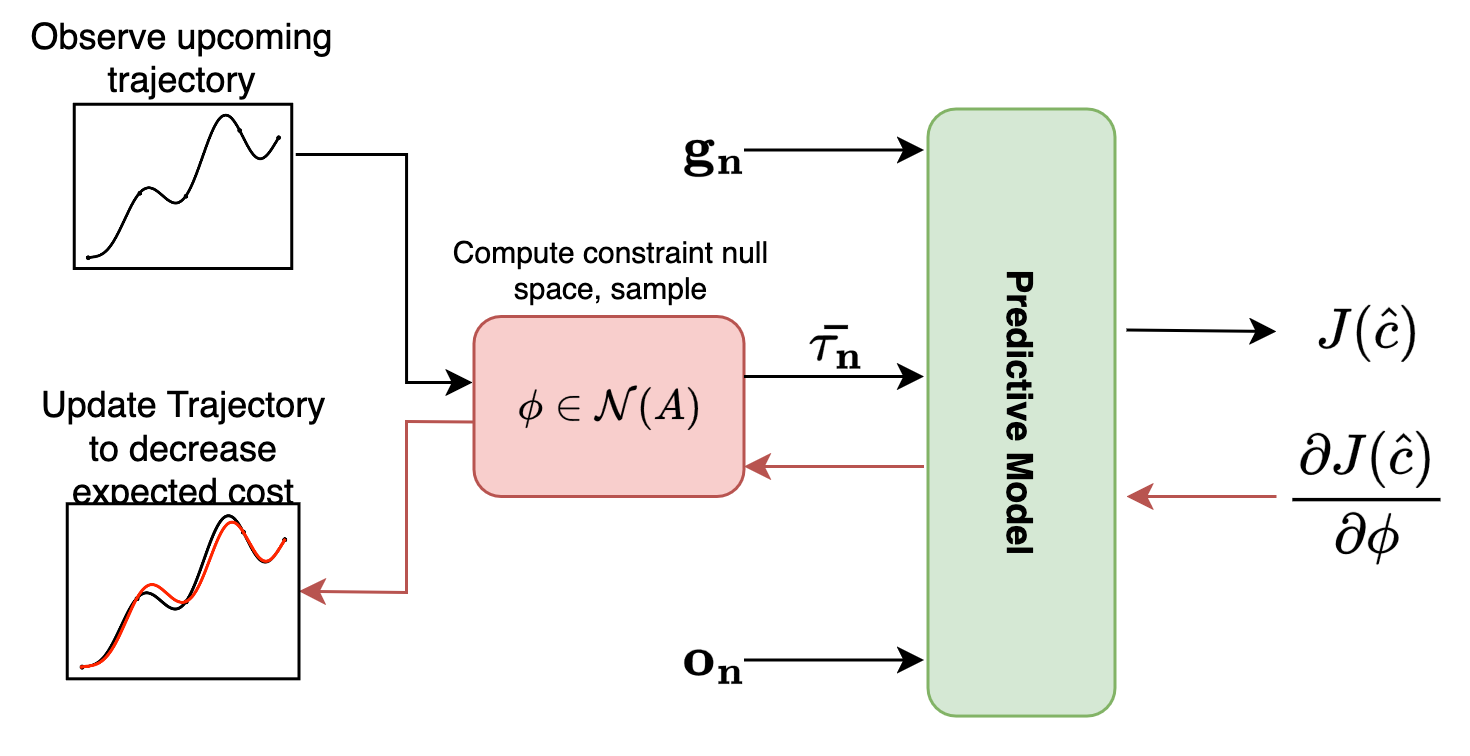}
    \caption{Trajectory adaptation using gradient backpropagation through the predictive model.}
    \label{fig:traj_opt}
\end{figure}

A minimum-snap (MinSnap) trajectory is the solution to this system of equations that minimizes the total snap (fourth derivative of position) over the course of the trajectory. 

An issue with the classical polynomial trajectory optimization formulation is that the flatness abstraction does not explicitly encode lower level actuator constraints which in practice may make the optimal trajectory dynamically infeasible. There are works that provide analytical \cite{muellerMotionPrimitive2015} and learned \cite{srikanthan2025quadlcdlayeredcontroldecomposition} functions relating spline coefficients to actuator constraints which can be used to further refine the trajectory. In this section, we demonstrate that the predictive model $\hat{F}$ can be used to improve the ``trackability" of a given MinSnap trajectory (or any given smooth, polynomial trajectory) by using gradient backpropagation from the model's predicted cost (Figure \ref{fig:traj_opt}). Because we encode the trajectory input to the model as samples from the trajectory, the main issue is how we can use the gradients from the model to update the trajectory in such a way that still satisfies the original smoothness and position constraints. 

Given any solution to $A\sigma = b$, if $\phi \in \text{Null}(A)$ then $A(\sigma + \phi) = b$. Thus, $\phi$ can be used to modify the original solution $\sigma$ while remaining within the solution space. We compute a basis for the nullspace of $A$ using SVD and denote the basis matrix as $V$, and denote the vector $\phi$ represented in the nullspace basis as $\phi_v$. To obtain the trajectory position at time $t$, $\tau(t)$, we first select the relevant spline segment by finding $t_i < t < t_{i+1}$ and create a selection matrix $S$ which is 1 at indices corresponding to active spline segment coefficients $\sigma_i^x, \sigma_i^y, \sigma_i^z$ and 0 elsewhere. Then, denoting the monomial vector as $\alpha = \left[(t-t_i)^{j}\right]_{j=0:d}$, $\bar{\tau}(t) = \alpha S[V\phi_v + \sigma]$. This technique is similar to null-space control in robotic manipulation \cite{hollerbach_null_space}. 

This function that computes the dense trajectory samples $\bar{\tau}[n:n+H]$ from the null coefficients $\phi_v$ is linear and therefore differentiable, allowing gradient backpropagation to update $\phi_v$. Given the current control gains $g_n$ and current state measurement $o_n$, we obtain the predicted cost from our model: $ \hat{c} = \hat{F}(g, o, \bar{\tau})$. Using autodifferentiation, we can compute $\frac{\partial \hat{c}}{\partial \phi_v}$, and use that to update $\phi_v \leftarrow \phi_v - \eta \frac{\partial \hat{c}}{\partial \phi_v}$ to find a new trajectory with lower predicted cost. By construction, the resulting trajectory will always satisfy the constraints after every gradient step without having to project the updated coefficients onto the feasible set. Besides optimization of existing trajectories, we note that the null-space of the constraint matrix can also be used on its own to generate random satisfying trajectories by sampling random vectors $\phi_v$, which can be useful for generating diverse tasks for model or policy training.

\section{Dataset Generation and Model Training}
\begin{figure}[t]
    \centering
    \includegraphics[width=0.4\textwidth]{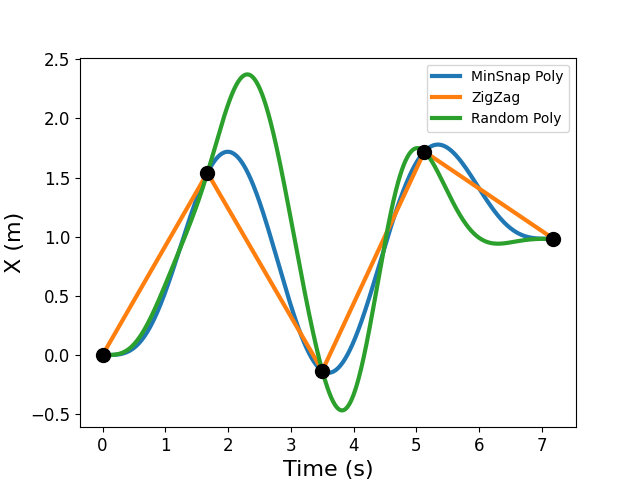}
    \caption{2D examples of the trajectory types included in our training dataset, interpolating between the same keypoints with the same average velocity.}
    \label{fig:traj_types}
\end{figure}

The performance prediction function $\hat{F}$ is modeled as a multi-layered perceptron (MLP) in PyTorch \cite{pytorch} with four hidden layers of sizes $[512, 512, 256, 256]$. We generate $8,000,000$ datapoints to train our model. Each datapoint is a tuple consisting of $(g_i, o_i, \bar{\tau}_i, c_i)$. The trajectories $\tau_i$ are generated by first randomly generating a sequence of three-dimensional waypoints in space, picking a random average velocity between (0.5, 3.0) m/s, then picking between three different possible trajectory types to interpolate between the points:
\begin{enumerate}
    \item MinSnap polynomial trajectories with constant average velocity between keypoints
    \item MinSnap polynomial trajectories with varying velocity between keypoints
    \item Random polynomial trajectories (not necessarily MinSnap) generated using the technique in Section \ref{sec:traj_adaptation}.
    \item ZigZag trajectories \cite{huang2023datt}, where the trajectory is a straight line between each of the waypoints (thus, there is a velocity discontinuity at each waypoint). 
\end{enumerate}

Examples of MinSnap, ZigZag, and Random Polynomial trajectories interpolating between the same waypoints in 2D are shown in Figure \ref{fig:traj_types}. These trajectory types were chosen with the aim of generalizing to arbitrary continuous or discontinuous trajectories. We specifically focus on polynomial spline trajectories due to their extensive use in trajectory planning for quadrotors, but also include the ZigZag trajectories to increase the generalizability of our model. 

Then, at a random time along the trajectory $t$, the initial position and velocity offsets from the trajectory at $t$ are randomly chosen as the initial state: $o_i= \tau(t) + \epsilon$. Finally, we pick a random $g_i$ and roll out the simulation for $H$ steps to obtain the performance, $c_i$. The quadrotor parameters used for all experiments are based on the CrazyFlie 2.0 platform \cite{CrazyflieBitcraze}. For all experiments, we use a simulation timestep of $\delta t = 0.01$s and a horizon $H=100$ steps, which corresponds to 1 second. The trajectory discretization timestep $\Delta T = 0.05$s.

\subsection{Parallelized Simulation}
We build upon RotorPy \cite{folk2023rotorpy}, a Python-based aerial robotics simulator, to collect simulation data to train our performance measure predictive models. RotorPy is chosen here because unlike many other simulators it offers higher fidelity aerodynamic models as well as explicit modeling of the motor dynamics and saturation constraints. To facilitate faster data collection, we refactored the multirotor dynamics, previously written in NumPy and SciPy, in PyTorch \cite{pytorch} using vectorized operations to allow for parallel simulation of a batch of quadrotors, following other massively parallelized simulators such as Isaac Sim \cite{NVIDIA_Isaac_Sim} and MuJoCo XLA \cite{todorov2012mujoco}. To take full advantage of the parallelized simulation, we also implemented vectorized versions of the geometric controller and trajectory generators to compute control commands and reference waypoints for a batch of simulated quadrotors in parallel. The vectorized simulation achieves a speedup of about 25x on CPU and also supports simulation on GPU, which can further improve performance for larger batch sizes. On an AMD Ryzen 3900X CPU, we obtain up to 17,000 FPS for a batch of 10,000 quadrotors. Leveraging multithreading on top of the vectorized simulation, we are able to generate the full dataset in about 7 hours. The vectorized implementations of the quadrotor dynamics, trajectory generators, and sensors have been open sourced for use in other applications such as reinforcement learning\footnote{Note to Reviewers: if accepted, a link to the code will be added here in the final version.}.

\section{Controller Optimization Experiments}
\begin{table*}[h]
% \tiny
\vspace{0.5em}
\centering
\caption{Average Tracking Error Results over 50 trajectories for each test case. Where relevant, failed runs are filtered out and the number of failures is reported in parentheses.}
\label{tab:gain_results}
\centering
\resizebox{0.8\linewidth}{!}{%
\begin{tabular}{@{}lccccc@{}}
\toprule
                & \multicolumn{1}{c}{\textbf{MinSnap}} & \multicolumn{1}{c}{\textbf{MinSnap-Hard}} & \multicolumn{1}{c}{\textbf{MinSnap-Varying}} & \multicolumn{1}{c}{\textbf{ZigZag}} &
                \multicolumn{1}{c}{\textbf{Lissajous}}\\ 
                \midrule
\textbf{Method} & \multicolumn{5}{c}{Avg Error, meters (\# crashes in parentheses), Lower is Better}  \\ \midrule
Nominal        & 0.44 (5)   & 0.8 (11) &  0.62 (6) & 0.47 (1) & 0.17 \\
Oracle (Static)  & 0.20 & 0.40 (1) & 0.25 & 0.33 & 0.12 \\ 
TACO (No-Traj)  & 0.38  & 0.66 (2) &  0.50 & 0.44 & 0.19 \\
\textbf{TACO (Full)} & \textbf{0.18}  & \textbf{0.30 (1)}     & \textbf{0.20} & \textbf{0.28} & \textbf{0.11} \\ \midrule
Oracle (Adaptive)   & 0.10  & 0.24      & 0.13 & 0.26 & 0.08 \\
MPC \cite{data_driven_mpc} & 0.19 & 0.21 (3) & 0.20 & 0.13 & 0.14 \\ \bottomrule
\end{tabular}}
\end{table*}

To measure how well our method optimizes the control gains for trajectory tracking, we evaluate it in a receding-horizon fashion (Figure \ref{fig:gain_opt}) on a simulated CrazyFlie tracking trajectories ranging from 6 to 10 seconds in length. At the beginning of each evaluation, the quadrotor begins at a hover at the first trajectory waypoint at $t=0$. Every $T=50$ timesteps (corresponding to 0.5s, or 2Hz), we use our method to optimize the control gains given the current state of the quadrotor $o_n$ and give the model the upcoming 1s segment of the reference trajectory. We measure the average position error obtained by the controller over the course of the entire trajectory.

We evaluate TACO and baselines on five different trajectory types: 50 MinSnap trajectories with constant average velocity between 1 m/s and 2.5 m/s, 50 MinSnap trajectories with constant average velocity between 1.5 m/s and 3.5 m/s (Minsnap-Hard), 50 MinSnap trajectories with a changing speed between each waypoint (MinSnap-Varying), 50 ZigZag trajectories, and 50 Lissajous curve trajectories, which are out-of-distribution for the model.

\subsection{Gain Optimization Baselines}
We benchmark our method against four baselines and one ablation.
\subsubsection{Nominal}
Our first baseline, the \texttt{Nominal} gains, are constant control parameters that were previously hand-tuned by a practitioner to a satisfactory level using step response testing. This most closely represent how gains are currently tuned: offline and based on a heuristic.
\subsubsection{Oracle (Static)}
Our second baseline is a Bayesian Optimization (BO) tuning method with access to the ground-truth simulation, which removes any implicit biases induced from hand-tuning. Specifically, we use BO to optimize the same parameters $g$ of the geometric controller. In this variant, the BO is allowed to query the true simulation of each trajectory, but is only allowed to choose \textit{one} gain to deploy for each trajectory. It optimizes the same weighted sum of performance measures as our method, but without the terminal cost. Because it has access to the ground truth cost function but cannot re-optimize the gains over the course of the trajectory, we call this baseline \texttt{Oracle (Static)}.
\subsubsection{Oracle (Adaptive)} We also compare our method to a BO oracle that is allowed to re-tune the control gains every $T\delta t=0.5s$, (with the same cost function as our method, including the terminal cost), referred to as \texttt{Oracle (Adaptive)}. This oracle is allowed access to ground truth simulation rollouts of the closed-loop dynamics over the upcoming horizon ($1$s, the same as \texttt{TACO}). We do not expect TACO to outperform \texttt{Oracle (Adaptive)} as the oracle has access to the ground truth simulation, but it provides an lower bound on the tracking error achievable by the geometric controller using trajectory-aware re-tuning.

\subsubsection{MPC} To contextualize the performance of the geometric control methods, we also evaluate the MPC controller from \cite{data_driven_mpc} without the aerodynamic adaptation module on our trajectory test sets. We use the default cost weights from \cite{data_driven_mpc}, and the MPC also is given a 1-second lookahead horizon. However, the MPC is allowed to re-plan after every control iteration. Although MPC is a different controller morphology, we include it as, similarly to our method, MPC approximates the solution to a receding-horizon optimal control problem. While MPC directly computes the individual motor thrusts and has access to a dynamics model, our method is constrained to optimizing the gains of the geometric controller and only uses the learned performance prediction function.

\subsubsection{TACO (No-Traj)} To confirm the importance of the trajectory lookahead, we also evaluate an ablation of our method, \texttt{TACO (No-Traj)}, which is trained on the same dataset but does not receive trajectory lookahead $\bar{\tau}$ as input. Instead, it must predict performance only based on the current quadrotor state. For this ablation, we add the current position error $e_p$ to the current state observation $o_n$. This ablation mimics to the quadrotor predictive model in \cite{pmlr-v270-sanghvi25a}, without meta learning and last-layer adaptation.
\begin{figure}[h]
    \centering
    \includegraphics[width=0.48\textwidth]{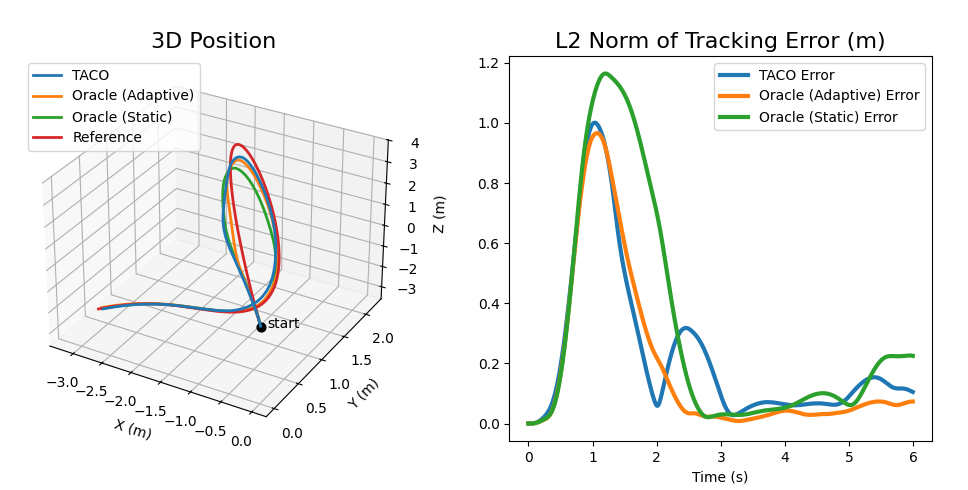}
    \caption{An example of tracking a challenging MinSnap trajectory using re-tuning with TACO (blue) versus the static oracle-tuned parameters (green) and adaptive oracle (orange). By retuning the control parameters online, TACO handles diverse maneuvers such going from a loop in altitude from $t=0:2$ to a gentle curve from $t=3:7$.}
    \label{fig:tracking_vv_eg}
\end{figure}

\subsection{Gain Optimization Results}

\begin{figure*}[t] % use * for spanning both columns
    \centering
    \subfloat[Adaptation for $t=0$ to $t=1$ \label{fig:traj_adapt_1}]{
        \includegraphics[width=0.24\textwidth]{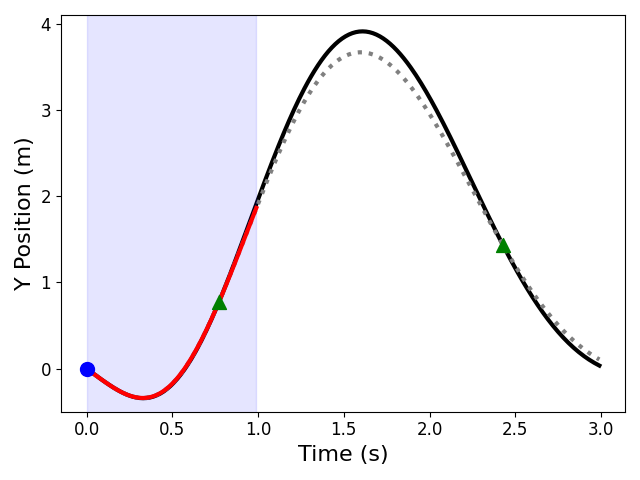}
    }
    \subfloat[Adaptation for $t=1$ to $t=2$\label{fig:traj_adapt_2}]{
        \includegraphics[width=0.24\textwidth]{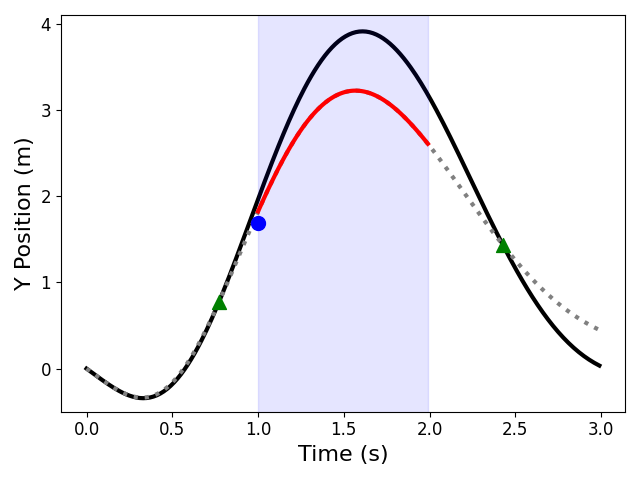}
    }
    \subfloat[Adaptation for $t=2$ to $t=3$\label{fig:traj_adapt_3}]{
        \includegraphics[width=0.24\textwidth]{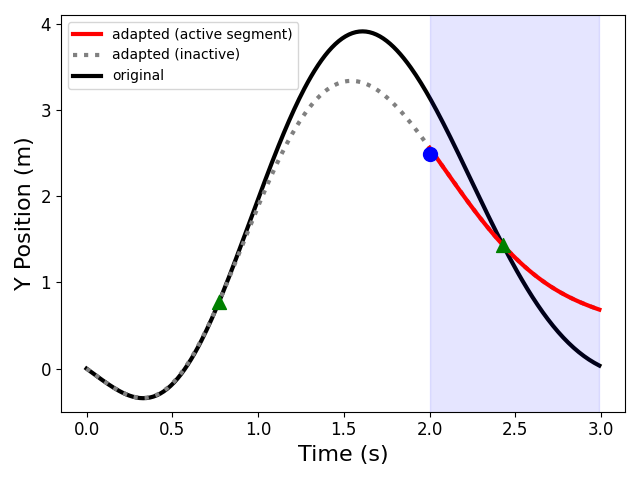}
    }
    \subfloat[Adapted trajectory (composed of segments in red) vs. original trajectory (black) \label{fig:traj_adapt_4}]{
        \includegraphics[width=0.2\textwidth]{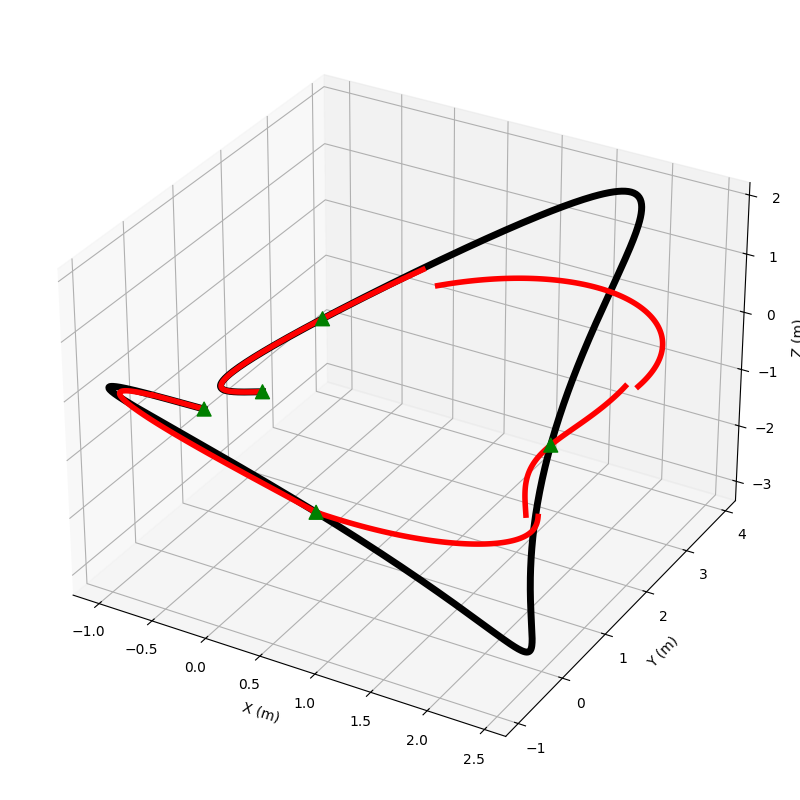}
    }
    \caption{Subfigures (a)-(c) show the sequence of re-adapted trajectories at $t=0, 1, 2$ in the $Y$ axis compared to the original trajectory in black. The highlighted portion shows the lookahead horizon of the model. The blue dot indicates the quadrotor's position at the beginning of each adaptation step. (d) shows the composite trajectory made up of the individual segments. While each adapted trajectory will be smooth, there may be a discontinuity at the drone's current position. However, each adapted trajectory still satisfies the time and waypoint constraints, shown by the fact that each adapted trajectory still passes through the keypoints, shown in green.}
    \label{fig:traj_adaptation}
\end{figure*}

We report the average tracking error on all trajectory test sets in Table \ref{tab:gain_results}, with failures filtered out and reported in parentheses. Failures are defined as rollouts where a method obtained tracking error $>$5m. On all trajectory test sets, TACO achieves the closest performance to \texttt{Oracle (Adaptive)}. The performance improvement over the \texttt{Nominal} baseline demonstrates the value of automated, task-specific controller optimization, while the improvement over \texttt{Oracle (Static)} demonstrates that re-optimizing the gains online further improves performance. The strong performance improvement of our full method over the \texttt{TACO (No Traj)} ablation also shows the value of including trajectory lookahead when tuning the gains. Furthermore, our strong results on the \texttt{Lissajous} test set, which was a trajectory type not seen during training, show that our training data captures sufficient variation to enable some out-of-distribution generalization.

We show an example of TACO and oracles tracking a MinSnap trajectory in Figure \ref{fig:tracking_vv_eg}, noting that our method tends to outperform the static oracle when more diverse, agile motions are required. Generally, the gap between the adaptive methods and the static methods grows as the trajectories grow increasingly challenging, as shown in the gap in performance in the ``MinSnap" and ``MinSnap-Hard" test cases. 

Both oracle methods are strong baselines; they have access to the ground truth cost function and are designed to run offline. Although \texttt{Oracle (Static)} only gets to select a single set of controller parameters for the entire trajectory, those parameters are still optimized with respect to the given trajectory and thus still achieve some level of task-specificity. However, the oracles take multiple minutes on a workstation to optimize the control gains at each optimization iteration, making them infeasible for online use on a robot. For example, with 10 parallel threads on a server-grade CPU, \texttt{Oracle (Static)} still takes approximately 45 minutes to optimize the gains for 50 trajectories. Meanwhile, our method is able to perform controller optimization in a fraction of a second, enabling online usage, and achieves superior performance to both \texttt{Nominal} and \texttt{Oracle (Static)}.

\begin{table}
\centering
\caption{Tracking Error (m) achieved in cross-comparisons of control gains tuned by TACO on different trajectories over 1s horizon}
\label{tab:taco_crossval}
\resizebox{0.4\textwidth}{!}{%
\begin{tabular}{lllllll}
    &                          & \multicolumn{4}{l}{\;\;\;\;\;\;\;\;\;Tested on} \\ \cline{3-6}
\multirow{5}{*}{\rotatebox[origin=c]{90}{Tuned For}} &                          & Slow       & Med        & Fast      & X-Fast   \\ \cline{3-6} 
    & \multicolumn{1}{l|}{Slow} & \textbf{0.016}     & 0.097      & 0.42      & 0.7  \\
    & \multicolumn{1}{l|}{Med}   & 0.033     & \textbf{0.066}     & 0.34      & 0.57  \\
    & \multicolumn{1}{l|}{Fast} & 0.086     & 0.14     & \textbf{0.2}     & \textbf{0.31}  \\
    & \multicolumn{1}{l|}{X-Fast}   & 0.12     & 0.19      & 0.26      & 0.34 
\end{tabular}}
\end{table}

Given the high dimensionality and covariances between the optimal gains, initial states, and trajectories, it is challenging to find a simple interpretation of the tuned gains in an online setting. However, we run the same experiment as in Table \ref{tab:optuna_crossval} and confirm that our method finds unique control gains for different trajectories, shown in Table \ref{tab:taco_crossval}, with only a slight error with the X-Fast trajectory, whose velocity lies at the edge of the model's training distribution.

\subsection{Gain Optimization on Real CrazyFlie}
To further validate our method, we use it to tune the gains of a PD controller on a real CrazyFlie 2.0, and compare it to a hand-tuned controller. The controller for the physical Crazyflie runs on a Robot Operating System (ROS) node running on a laptop with an Apple M3 processor. The ROS node receives the quadrotor's position through a Vicon motion-capture system\footnote{https://www.vicon.com/} running at 100Hz and computes a desired collective thrust and quaternion attitude command for the quadrotor, which the onboard PID controller then converts into motor speeds. We run TACO on the same laptop synchronously with the controller, with a replanning frequency of 2Hz, which does not interrupt the control loop severely as each TACO iteration takes $\sim$20ms to execute.

We find that our method is able to reduce the average tracking error by \textbf{5cm} over the course of the trajectory compared to the hand-tuned control parameters. Much of this comes from improved $z$-axis tracking while maintaining similar performance in the $x$ and $y$ axes. This is a difficult task for TACO due to a significant gap between simulation and reality: the real quadrotor is less capable of agile maneuvers than the simulated one, and its attitude dynamics are significantly different from simulation. Furthermore, we use the same model used for the simulation experiments, which was trained \textbf{without} any sim2real techniques (e.g. domain randomization \cite{pengSimtoRealTransferRobotic2018a}). 

\section{Trajectory Adaptation Experiments}

We also test our online trajectory adaptation framework on thirty trajectories in simulation. Each evaluation begins with a pre-planned MinSnap trajectory interpolating between 5 keypoints which the quadrotor must track. Every $H=1$ seconds, the model is asked to re-optimize the upcoming segment of the reference trajectory, as shown in Figure \ref{fig:traj_adaptation}. The number of gradient steps taken per adaptation iteration and the step size $\eta$ are kept the same for every trajectory. We measure the quadrotor's tracking error at each of the constraint keypoints and compare that to the error obtained if the original trajectory had been followed. If we fix $g$ to its nominal values, the quadrotor obtains an average keypoint error of \textbf{0.21m} when tracking the online-adapted trajectories compared to \textbf{0.31m} with the static nominal trajectories, showing an improvement of \textbf{32\%}. Performing online gain optimization alongside trajectory adaptation yields further performance improvements, decreasing the average keypoint tracking error to \textbf{0.15m}.

In Figure \ref{fig:traj_adaptation}, we show an example of a sequence of optimized trajectories, compared to the nominal trajectory. Each optimized trajectory satisfies the same smoothness and waypoint constraints as the nominal trajectory. Because the trajectory is being re-adapted at regular intervals, there are discontinuities in the composite trajectory where the newly adapted trajectory does not match the previously adapted trajectory. 

\section{Conclusion and Future Work}
In this work, we have presented a method, TACO, to optimize the controller parameters for a commonly-used quadrotor controller based on the upcoming trajectory segment and current quadrotor state. Our method surpasses both heuristic- and oracle-based static controller tuning, providing a step away from time-intensive, manual controller tuning towards automatic, task-specific adaptation. TACO is also computationally efficient enough to be used in real-time to optimize the control paramters on a real CrazyFlie, demonstrating a limited capability for zero-shot sim2real transfer. Furthermore, the predictive model can be used for online replanning to increase the dynamic feasibility of polynomial spline trajectories online while respecting smoothness and waypoint constraints.

The principle limitation of this work is that the predictive model was trained only on the parameters of a single quadrotor. One line of future work would be to incorporate methods for domain adaptation into TACO's predictive model, enabling gain tuning that accounts for both the upcoming trajectory and potentially varying quadrotor dynamics. Another limitation is that, while the gain optimization runs at real-time rates, each trajectory adaptation step (consisting of multiple gradient updates) takes $\sim$1 second to execute, which is slow when the horizon is itself 1 second; future work also includes speeding trajectory adaptation up enough for real-time usage. TACO is also not restricted to the geometric controller studied in this work; other future work includes applying TACO to different controller morphologies, such as MPC.

%    \begin{figure}[thpb]
%       \centering
%       \framebox{\parbox{3in}{We suggest that you use a text box to insert a graphic (which is ideally a 300 dpi TIFF or EPS file, with all fonts embedded) because, in an document, this method is somewhat more stable than directly inserting a picture.
% }}
%       %\includegraphics[scale=1.0]{figurefile}
%       \caption{Inductance of oscillation winding on amorphous
%        magnetic core versus DC bias magnetic field}
%       \label{figurelabel}
%    \end{figure}

% \addtolength{\textheight}{-12cm}   % This command serves to balance the column lengths
%                                   % on the last page of the document manually. It shortens
%                                   % the textheight of the last page by a suitable amount.
%                                   % This command does not take effect until the next page
%                                   % so it should come on the page before the last. Make
%                                   % sure that you do not shorten the textheight too much.
                                  
\bibliography{example}

\end{document}